\documentclass[conference,a4paper]{IEEEtran}
\IEEEoverridecommandlockouts

\usepackage[hidelinks]{hyperref}
\usepackage[cmex10]{amsmath}
\usepackage{amssymb} 
\usepackage{amsfonts}
\usepackage{marvosym}
\usepackage{pifont}  

\usepackage{dblfloatfix}

\usepackage[ruled,vlined]{algorithm2e}
\usepackage{graphicx}
\graphicspath{{Figures/PDF/}{Figures/PNG/}}

\usepackage{array} 
\usepackage{xcolor}
\usepackage{color}

\usepackage{colortbl} 

\usepackage{booktabs}
\usepackage{siunitx}
\usepackage[numbers,compress]{natbib}
\usepackage{texnames}
\usepackage{bm,bbm}
\usepackage{orcidlink}
\usepackage{multirow}

\begin{document}

\title{\ Mamba-MOC: A Multicategory Remote Object Counting via State Space Model
\thanks{This work was supported in part by the National Natural Science Foundation of China under Grant 62271418, and in part by the Natural Science Foundation of Sichuan Province under Grant 2023NSFSC0030 and 2025ZNSFSC1154, in part by the Postdoctoral Fellowship Program and China Postdoctoral Science Foundation under Grant Number BX20240291, and in part by the Fundamental Research Funds for the Central Universities under Grant 2682025CX033. \textit{(Corresponding author: Heng-Chao Li)}}
}

\author{	\IEEEauthorblockN{Peng Liu\orcidlink{0009-0005-2577-7371}}
	\IEEEauthorblockA{\textit{Southwest Jiaotong University}\\
		610031 Chendu, China\\
		lp@my.swjtedu.cn}
	
	\and
	\IEEEauthorblockN{Sen Lei\orcidlink{0000-0002-8010-3282}}
	\IEEEauthorblockA{\textit{Southwest Jiaotong University}\\
		610031 Chendu, China\\
		senlei@swjtu.edu.cn}
	
	\and
	\IEEEauthorblockN{Heng-Chao Li\orcidlink{0000-0002-9735-570X}}
	\IEEEauthorblockA{\textit{Southwest Jiaotong University}\\
		610031 Chendu, China\\
		lihengchao\_78@163.com}
}

\maketitle
\begin{abstract}
	Multicategory remote object counting is a fundamental task in computer vision, aimed at accurately estimating the number of objects of various categories in remote sensing images. Existing methods rely on CNNs and Transformers, but CNNs struggle to capture global dependencies, and Transformers are computationally expensive, which limits their effectiveness in remote applications. Recently, Mamba has emerged as a promising solution in the field of computer vision, offering a linear complexity for modeling global dependencies. To this end, we propose Mamba-MOC, a mamba-based network designed for multi-category remote object counting, which represents the first application of Mamba to remote sensing object counting. Specifically, we propose a cross-scale interaction module to facilitate the deep integration of hierarchical features. Then we design a context state space model to capture both global and local contextual information and provide local neighborhood information during the scan process. Experimental results in large-scale realistic scenarios demonstrate that our proposed method achieves state-of-the-art performance compared with some mainstream counting algorithms. The code will be available at \url{https://github.com/lp-094/Mamba-MOC}.
\end{abstract}

\begin{IEEEkeywords}
	Remote sensing, object counting, mamba, state-space model(SSM).
\end{IEEEkeywords}

\section{Introduction}

Object counting in remote sensing scenes, which involves estimating the number of objects of a specific category in given images, has become increasingly significant in applications such as urban planning\cite{rathore2016urban}, agriculture monitoring\cite{chen2022transformer}, and ecological survey \cite{sarwar2018detecting}. Compared to traditional counting tasks, this task presents greater challenges due to the broader spatial coverage and more complex scene content. 

Conventional methods for remote sensing object counting predominantly rely on Convolutional Neural Networks (CNNs) or Vision Transformers (ViTs). CNN-based approaches are widely utilized due to their translational invariance and linear computational complexity. However, the fixed-size local connections inherent in CNNs constrain their receptive fields, limiting the ability to effectively capture long-range dependencies and hindering the modeling of global context. In contrast, ViT-based approaches exploit long-range dependencies, which are important for visual tasks \cite{10816052} and dynamic weight allocation to enhance the modeling of visual information. However, the self-attention mechanism inherent in ViTs introduces quadratic complexity with respect to input size, which results in a significant increase in computational cost, especially when handling large-scale data.

To address the limitations, the Selective Structured State-Space Model (S6), or Mamba, incorporates selective scanning techniques and hardware-optimized designs, enabling effective modeling of long-range dependencies while maintaining linear complexity and dynamic weight allocation. This capability has garnered significant attention in visual tasks, with numerous studies exploring Mamba’s potential in areas such as image classification\cite{liu2024vmambavisualstatespace} and facial expression recognition \cite{ma2024fer}. However, its application in remote sensing counting tasks remains largely unexplored.

Building on this, we propose Mamba-MOC, a Mamba-based framework specifically designed for remote sensing object counting, aimed at exploring the potential of Mamba in this domain. Specifically, we utilize the visual Mamba model(Vmamba) as the backbone to extract more discriminative features. We also design a Mamba-based cross-scale interaction module to enable effective interaction between multi-level features, thereby improving the representation of remote sensing objects across different scales. Furthermore, we propose a context state space model that mitigates the inherent limitations of causal scanning in the Mamba model when applied to 2D images by incorporating a local convolution operation. This operation also extracts local contextual information, which is subsequently integrated into the global context. Consequently, the model enables a more effective extraction of fine-grained information and the experimental results validate its effectiveness.

\begin{figure*}[!t]
	\centering
	\includegraphics[width=7.in]{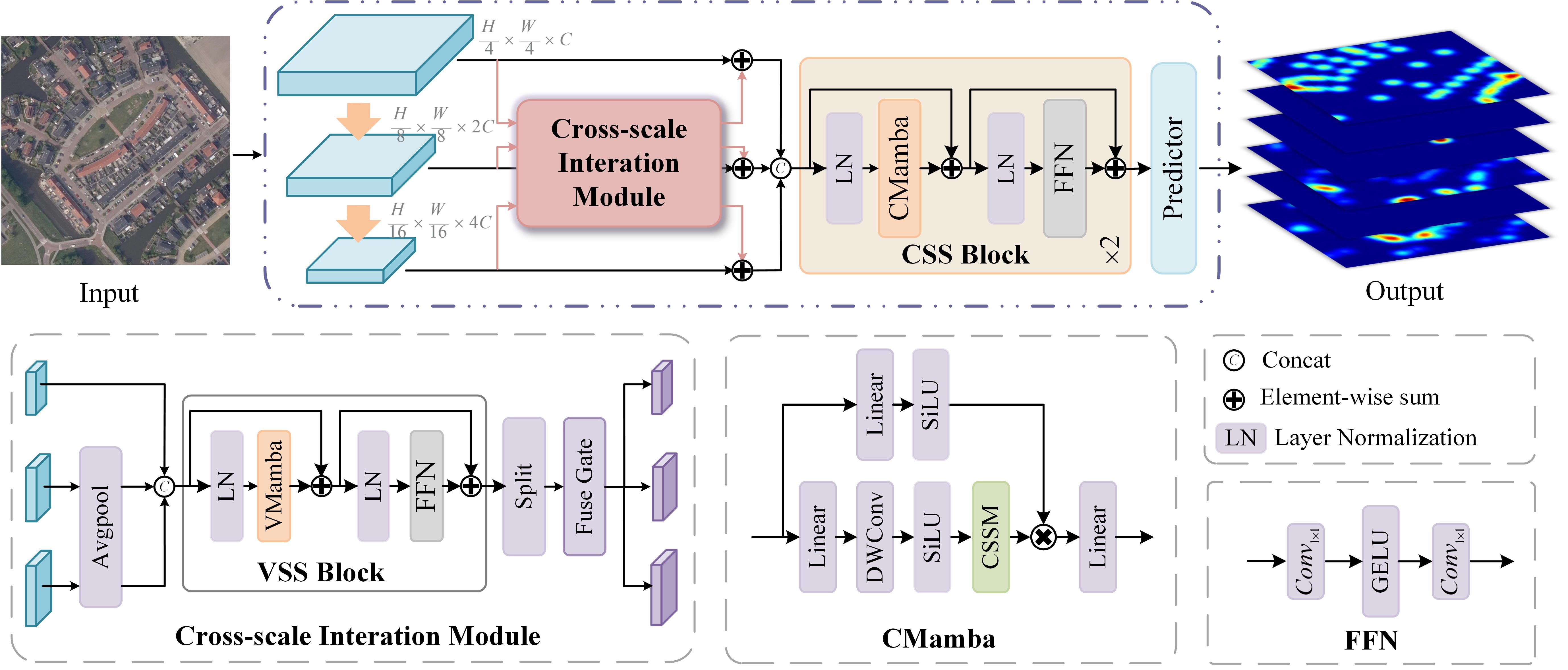}
	\caption{Overview of the proposed Mamba-MOC.}
	\label{fig_4}
\end{figure*}

\section{Preliminaries}
State Space Models (SSM) \cite{gu2023mamba}, inspired by continuous linear time-invariant (LTI) systems, map a one-dimensional input  $x(t)\in \mathbb{R}^{L}$ to the output $y(t)\in \mathbb{R}^{L}$ through a hidden state $h(t)\in \mathbb{R}^{L}$ with linear complexity, which are formally defined by the following mathematical formulation:
\begin{equation}
	\begin{split}
		\label{eq_1}
		h^{'}(t)=\mathbf{A} h(t) + \mathbf{B} x(t), y(t)=\mathbf{C} h(t)
	\end{split}
\end{equation}
where $\mathbf{A}\in \mathbb{R}^{N\times N} $, $\mathbf{B}\in \mathbb{R}^{N\times 1} $ and $\mathbf{C}\in \mathbb{R}^{N\times N} $ are the projection matrix, $N$ is the state size.
To adapt this to discrete data such as images and text, it is necessary to discretize the aforementioned continuous system. Mamba adopt zero-order hold with a timescale parameter $\triangle$ to transform the continuous parameters $\mathbf{A}$ and $\mathbf{B}$ from the continuous system into the discrete parameters $\overline{A} $ and $\overline{B} $:
\begin{equation}
	\begin{split}
		\label{eq_2}
		\overline{\mathbf{A}} = exp(\triangle \mathbf{A} ),\overline{B} = (\triangle \mathbf{A} )^{-1} (exp(\triangle \mathbf{A} )-I)\cdot\triangle\overline{B}
	\end{split}
\end{equation}
After that, the discretized model can be represented as:
\begin{equation}
	\begin{split}
		\label{eq_3}
		h_{t} =  \mathbf{\overline{A}} h_{t-1} + \mathbf{\overline{B}} x_{t}, y_{t}=\mathbf{C} h_{t}
	\end{split}
\end{equation}
Finally, it can be represented as a global convolution, formally defined as:
\begin{equation}
	\begin{split}
		\label{eq_4}
		\mathbf{\overline{K}} = (\mathbf{C}\mathbf{\overline{B}}, \mathbf{C}\mathbf{\overline{AB}},...,\mathbf{C}{\mathbf{\overline{A}}}^{L-1}  \mathbf{\overline{B}}),y=x\ast \overline{\mathbf{K}} 
	\end{split}
\end{equation}
where $\ast$ denotes convolution operation, $L$ is the length of input $x$, and $\overline{\mathbf{K}}\in \mathbb{R}^{L} $ is a structured convolutional kernel.

\section{Methodology}

\subsection{Overall Architecture}
As illustrated in Fig. 1, our proposed method mainly comprises a vmamba\cite{liu2024vmambavisualstatespace} backbone, a cross-scale interaction module, two Context State Space (CSS) block. The vmamba backbone is employed as the encoder to extract multi-level feature representations {$F_{b}^{1}\in \mathbb{R} ^{\frac{H}{4} \times \frac{H}{4}\times C}$, $F_{b}^{2}\in \mathbb{R} ^{\frac{H}{8}\times \frac{W}{8}\times 2C}$, $F_{b}^{3}\in \mathbb{R} ^{\frac{H}{16}\times \frac{W}{16}\times 4C}$}. The cross-scale interaction module plays a critical role in effectively integrating coarse-grained and fine-grained features, thus augmenting the interaction of information across multiple scales. And the CSS block is equipped with a Contextual State Space Model (CSSM), which is specifically designed to capture and refine contextual information and focus on local neighborhood details during scanning process. Ultimately, the final feature is forwarded to a predictor to generate the density prediction.  

\begin{table*}[]
	\centering
	\caption{Comparison with the state-of-the-art methods on NWPU-MOC datasets. The best results are highlighted in red.}
	\renewcommand\arraystretch {1.2}
	\definecolor{MyRed}{RGB}{255,0,0}
	\definecolor{MyBlue}{RGB}{0,0,255}
	\setlength\tabcolsep{0.8mm}%
	\label{table1}
	\begin{tabular}{l|lllllllllllllllllll|ll}
		\toprule[1pt]
		\multirow{2}{*}{Methods}    &    & \multicolumn{2}{c}{Ship}   &   & \multicolumn{2}{c}{Vehicle}   &    & \multicolumn{2}{c}{Building}  &  & \multicolumn{2}{c}{Countainer}  &  & \multicolumn{2}{c}{Tree}  &   & \multicolumn{2}{c}{Airplane}  & &\multirow{2}{*}{$\overline{\mathrm{MSE}}$ $\downarrow$} &\multirow{2}{*}{WMSE $\downarrow$}\\ \cline{3-4} \cline{6-7} \cline{9-10} \cline{12-13}  \cline{15-16} \cline{18-19}
		& & MAE $\downarrow$ & MSE $\downarrow$ &     & MAE $\downarrow$ & MSE $\downarrow$  &   & MAE $\downarrow$ & MSE $\downarrow$  &   & MAE $\downarrow$ & MSE $\downarrow$ &   & MAE $\downarrow$ & MSE $\downarrow$  &   & MAE $\downarrow$ & MSE $\downarrow$  & &  &  \\ \hline
		CSRNet \cite{li2018csrnet}   &  & 4.6024  & 17.4716  &  & 13.7151   & 24.3899 &   & 8.0835    & 14.8328  &   & 6.5580   & 25.4806  &   & 27.8693  & 47.7390  &   & 0.2623 & 0.5090  &  & 21.7371 & 124.1803 \\ 
		SFCN \cite{Wang2019sfcn}   &  & 1.7080  & 10.5572  &  & 5.5301   & 13.8747 &   & 4.0828    & 8.3646  &   & 1.8993   & 7.3178  &   & 15.8311  & 32.2336  &   & \color{MyRed}{0.0615} & \color{MyRed}{0.4768}  &  & 14.8041 & 77.6888 \\ 
		PSGCNet \cite{gao2022psgcnet}   &  & 3.7138  & 17.5604  &  & 10.1730   & 18.5158 &   & 9.8801    & 18.9321  &   & 5.8081   & 25.7689  &   & 21.2395  & 40.2752  &   & 0.1204 & 0.4765  &  & 20.2548 & 82.2757 \\ 
		DSACA \cite{xu2021dsaca}   &  & 1.5117  & 4.0399  &  & 6.2882   & 38.7013 &   & 3.6883    & 7.2121  &   & 3.5049   & 9.9215  &   & 17.3888  & 32.9396  &   & 0.1100 & 0.4757  &  & 15.5483 & 97.0073 \\ 
		MCC \cite{gao2024moc}   & & 0.9684   & 4.3597  &  & \color{MyRed}{4.1432}    & 12.7091 &   & 3.4963     & 7.3538  &   & 1.6867    & 6.8366   &   & 16.7922   & 35.3594   &   & \color{MyRed}{0.0615}  & \color{MyRed}{0.4768}   &  & 11.4826 & 43.1082 \\ \hline
		\rowcolor{gray!25} Ours       &   & \color{MyRed}{0.7373}   & \color{MyRed}{4.0277}  &  & 4.4146    & \color{MyRed}{10.5133}  &   & \color{MyRed}{2.9027}     & \color{MyRed}{6.4310} &   & \color{MyRed}{1.6213}   & \color{MyRed}{5.5722}  &   & \color{MyRed}{15.6003}  & \color{MyRed}{30.4554}  &   & \color{MyRed}{0.0615}  & \color{MyRed}{0.4768}  & & \color{MyRed}{9.5794} & \color{MyRed}{27.2012} \\ 
		\bottomrule[1pt]
	\end{tabular}
\end{table*}

\subsection{Cross-scale Interaction Module}
For the input image $I\in \mathbb{R} ^{H\times W\times 3}$, leveraging the Vmamba backbone, multi-scale features from coarse to fine are extracted to effectively address the scale variation challenges in aerial image. However, a critical issue remains in exploiting the interactions between the coarse and fine stages. To address this challenge, we propose a Cross-Scale Interaction Module (CIM) to enhance the interplay between the coarse and fine stages. 

\begin{figure}[!t]
	\centering
	\includegraphics[width=3. in]{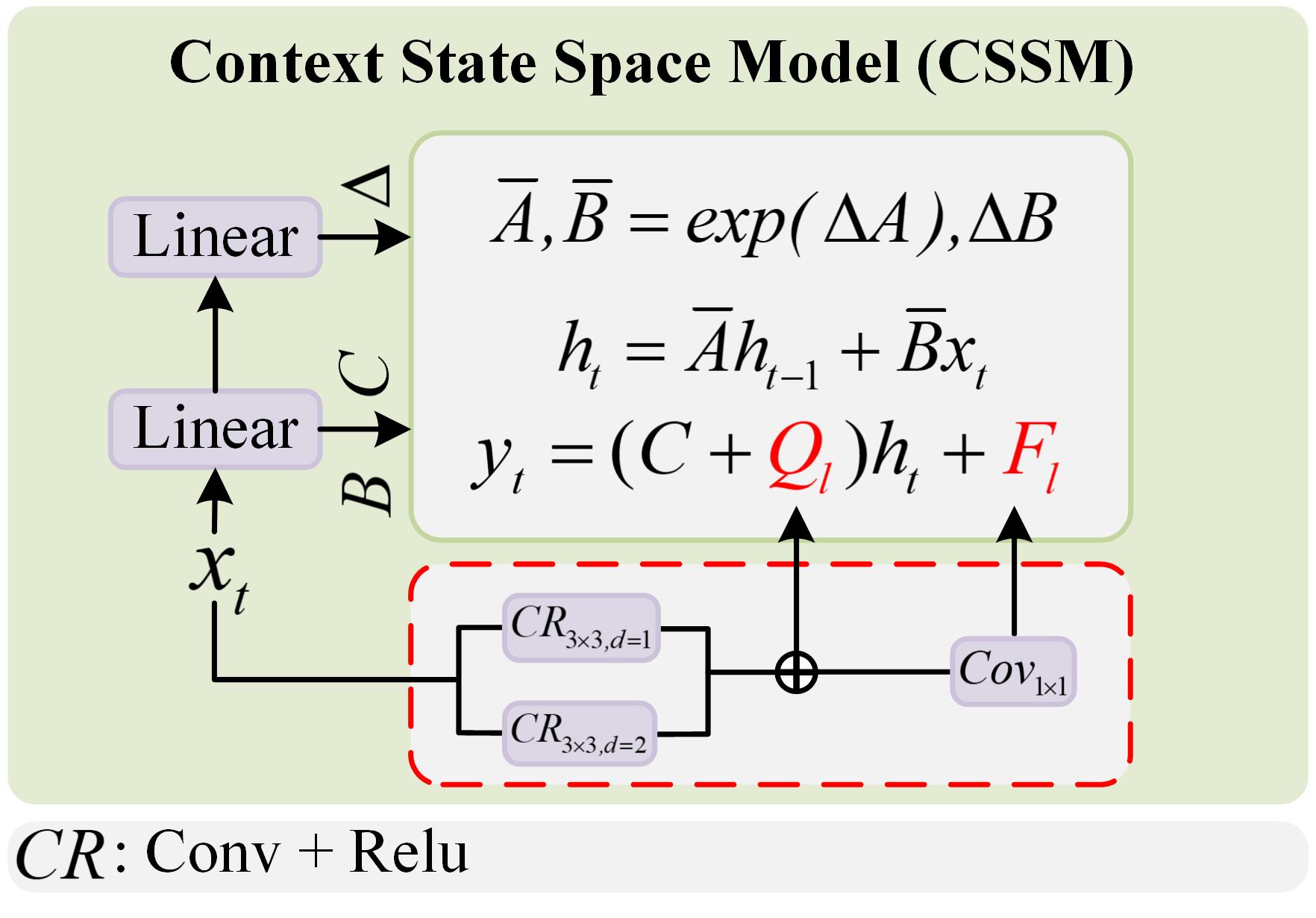}
	\caption{Architecture of the context state space model.}
	\label{fig_2}
\end{figure}

Specifically, for the extracted features $\left \{F_{b}^{1},F_{b}^{2},F_{b}^{3} \right \} $, we first align these features by resizing $F_{b}^{1}$ and $F_{b}^{2}$ to match the spatial shape of $F_{b}^{3}$, and subsequently concatenate them along the channel dimension, as illustrated below:
\begin{equation}
	\begin{split}
		\label{eq_10}
		F_{r}^{i} &= \mathrm{Avgpool} (F_{b}^{i}) , i\in \left \{ 1,2 \right \}  \\
		F_{align} &= Conv_{1\times 1}(\mathrm{Concat} (F_{r}^{1}, F_{r}^{2}, F_{b}^{3}))
	\end{split}
\end{equation}
where $F_{r}^{i}$ represents the pooled features, $Conv_{1\times 1}$ is  $1\times 1$ convolution and $F_{align}\in \mathbb{R} ^{\frac{H}{16}\times \frac{W}{16}\times 2C}$ represents the features obtained by concatenating multi-stage features along the channel dimension, followed by the linear projection.

Then, the aligned features are fed into the VSS block, which is the core component of Vmamba, for self-fusion and mapping back to the original channel dimensions through $1\times 1$ convolution. 
\begin{equation}
	\begin{split}
		\label{eq_11a}
		F_{fuse} &= Conv_{1\times 1}(\mathrm{Mamba} (F_{align})) 
	\end{split}
\end{equation}

After that, We split $F_{fuse}$ into three parts along the original channel dimensions, each of which is upsampled to match the size of $F_{b}^{i}$. The gating mechanism from \cite{liu2024rotated} is then used to integrate these parts with $F_{b}^{i}$, effectively   mitigating the semantic discrepancy among them.

\subsection{Context State Space Model}

Inspired by \cite{guo2024mambairv2}, the output matrix $\mathbf{C}$ in Equation 1 functions as a query in the attention mechanism. This analysis yields two critical findings: (1) Through local 2D modeling, $\mathbf{C}$ effectively ``queries'' pixels within the local context, addressing the inherent limitations of causal scanning in the Mamba model when applied to 2D images. (2) By emphasizing local context, the model focuses on fine-grained details while simultaneously integrating global context through global scanning, which improves the extraction of target features.

\begin{figure}[!t]
	\centering
	\includegraphics[width=3.5in]{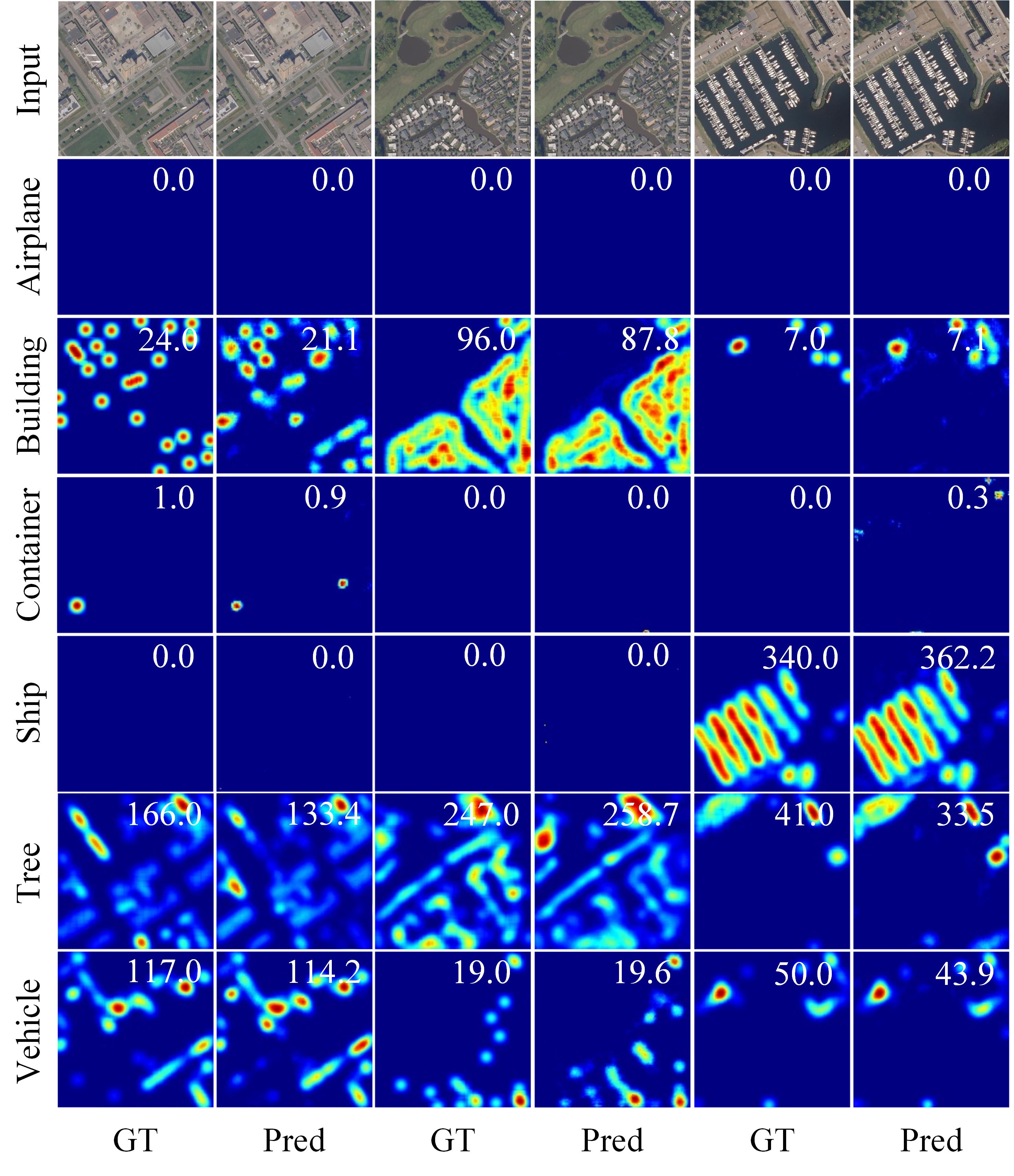}
	\caption{Visualization results of Mamba-MOC on NWPU-MOC dataset.}
	\label{fig_3}
\end{figure}

Based on this concept, we propose a Context State Space Model (CSSM) in a straightforward manner. As shown in Fig. \ref{fig_3}, the proposed CSSM incorporates multi-scale local information, allowing the model to learn and capture correlations between local neighbourhoods within the image. This method implicitly mitigates the inherent limitations of the SSM, which compromise the preservation of spatial structure. At the same time, it integrates both local and global contextual information, improving the perception of counting targets. Specifically, given an input feature map $F_{c}\in \mathbb{R} ^{H\times W\times C}$, we first capture multi-scale contextual features $F_{ms}\in \mathbb{R} ^{H\times W\times D}$ by applying multiple convolutional operations, from which local query $Q_l\in \mathbb{R} ^{HW\times D}$ is obtained through a flattening step. Simultaneously, a $1\times 1$ convolution is applied for linear projection to yield local feature $F_l\in \mathbb{R} ^{H\times W\times C}$, as expressed in the following equation:
\begin{equation}
	\begin{split}
		\label{eq_12}
		F_{ms} &= CR_{3\times 3,d=1}(F_c) + CR_{3\times 3,d=2}(F_c), \\
		Q_l &= Flatten(F_{ms}), \\
		F_l &= Conv_{1\times 1}(F_{ms})
	\end{split}
\end{equation}
where $CR_{3\times 3,d=1}$ and $CR_{3\times 3,d=2}$ represent $3\times 3$ convolution with a dilation rate of 1 and 2, respectively, followed by a relu activation function, and $Conv_{1\times 1}$ is $1\times 1$ convolution.

Then, we incorporate $Q_l$ into $C$ throgh residual addition and incorporate $F_l$ into $y(t)$ to formulate the context state space model:
\begin{equation}
	\begin{split}
		\label{eq_13a}
		h^{'}(t)=\mathbf{A} h(t) + \mathbf{B} x(t), \\
		y(t)=(\mathbf{C} + \mathbf{Q_l}) h(t) + \mathbf{F_l}
	\end{split}
\end{equation}

Through the aforementioned design, We adopt CSSM as the core component and, following the structure outlined in \cite{liu2024vmambavisualstatespace}, construct our CMamba, which is then used to build the CSS Block, as illustrated in Fig. 1. This allows the integration of global-local contextual information while paying attention to local neighborhood information during the scanning process.

\section{Experimental Results}

\subsection{Related Settings}
\textbf{Datasets:} We have conducted comprehensive experiments on the recently released multi-object remote sensing counting dataset, NWPU-MOC\cite{gao2024moc}. This dataset is derived from 3,416 aerial and remote sensing scenes, containing a total of 383,195 annotated points across 14 categories. It consists of 3,416 images, where the training and testing sets partitioned according to the official distribution. In particular, 2,391 images are allocated to the training set, and the remaining 1,025 images are used for testing. In our experiments, following \cite{gao2024moc}, we also roughly group the NWPU-MOC dataset into 6 categories.

\textbf{Evaluation Metrics:} The performance of the proposed framework is assessed using four metrics: mean absolute error(MAE), root mean squared error(RMSE), intercategory average MSE($\overline{\mathrm{MSE}}$) and weighted MSE(WMSE). $\overline{\mathrm{MSE}}$ and WMSE are proposed by \cite{gao2024moc}.

\begin{table}[]
	\centering
	\caption{Ablation studies of each component in our method. Bold font indices the best performance}
	\renewcommand\arraystretch {1.3}
	\setlength\tabcolsep{5mm}%
	\label{table2}
	\begin{tabular}{llllll}
		\toprule[1pt]
		Method &  $\overline{\mathrm{MSE}}$ $\downarrow$   &WMSE $\downarrow$   \\ \hline
		Baseline&      10.3800 & 34.5672 \\
		Baseline+CIM&      10.3533 & 33.8590 \\
		Baseline+CIM+CSSM&      \textbf{9.5794} & \textbf{27.2012} \\ \bottomrule[1pt]
	\end{tabular}
\end{table}

\textbf{Implementation Details:}
The ground truth density maps for the experiments are generated using a Gaussian kernel with a bandwidth of 4 and a size of 15. All experiments are performed within the PyTorch framework and conducted on an NVIDIA RTX 4090 GPU. During training, the input resolution is set to $512\times512$, and the network is optimized using AdamW optimizer with a learning rate of 5e-5, weight decay of 1e-4, and a batch size of 8, with a total of 200 epochs.

\subsection{Comparison with state-of-the-arts}
To provide a comprehensive benchmark evaluation, we compare our method with several state-of-the-art counting methods on the NWPU-MOC dataset. The results are summarized in Tabel \ref{table1}. It show that our approach outperforms existing methods in terms of $\overline{\mathrm{MSE}}$ and WMSE, reducing them to 9.5794 and 27.2012, respectively. In category-level analysis, our method achieves the best performance in five out of six categories, with only a 0.3 higher MAE in the Vehicle category compared to the best-performing method. Overall, the experimental findings validate the effectiveness of our approach.

The visual results of Mamba-MOC on the NWPU-MOC dataset are presented in Fig. 3. Three images from different scenes in the dataset are selected. As shown, the counting values generated by our method are highly accurate and closely match the ground truth counts. The results highlight our model’s ability to effectively handle diverse scenarios and provide precise crowd estimations across varying conditions.

\subsection{Ablation study}
To evaluate the effectiveness of CIM and CSSM, ablation experiments are conducted on the NWPU-MOC dataset, with a network incorporating a backbone and FPN structure as the baseline. As shown in Table \ref{table2}, the addition of CIM results in a reduction in both MSE and WMSE, which implies that the inter-scale interactions facilitated by Mamba play a crucial role in promoting effective feature fusion across different scales. When CSSM is further integrated, the $\overline{\mathrm{MSE}}$ and WMSE are reduced to 9.5794 and 27.2012, highlighting the significant contribution of contextual information in enhancing the extraction of relevant features from the fused representations. These results validate the effectiveness of the proposed method for multicategory remote object counting task.

\section{Conclusion}
In this paper, we introduce Mamba-MOC, a Mamba-based method for remote sensing multi-object counting. Initially, we leverage the advantages of Mamba's global modeling capabilities to design a cross-scale module that effectively integrates multi-scale and multi-granularity features within the FPN structure. Additionally, we propose a contextual state space model that overcomes the limitation of Mamba's causal scanning process, which struggles to capture local neighborhood context. At the same time, this model integrates both local and global contexts, enhancing the network's ability to interpret counting targets more effectively. Experimental results demonstrate the promising performance of our method in remote sensing multi-object counting tasks and validate the effectiveness of the Mamba framework in this domain.

\small
\bibliographystyle{IEEEtranN}
\bibliography{Mamba-MOC}

\begin{thebibliography}{14}
\providecommand{\natexlab}[1]{#1}
\providecommand{\url}[1]{#1}
\csname url@samestyle\endcsname
\providecommand{\newblock}{\relax}
\providecommand{\bibinfo}[2]{#2}
\providecommand{\BIBentrySTDinterwordspacing}{\spaceskip=0pt\relax}
\providecommand{\BIBentryALTinterwordstretchfactor}{4}
\providecommand{\BIBentryALTinterwordspacing}{\spaceskip=\fontdimen2\font plus
\BIBentryALTinterwordstretchfactor\fontdimen3\font minus \fontdimen4\font\relax}
\providecommand{\BIBforeignlanguage}[2]{{%
\expandafter\ifx\csname l@#1\endcsname\relax
\typeout{** WARNING: IEEEtranN.bst: No hyphenation pattern has been}%
\typeout{** loaded for the language `#1'. Using the pattern for}%
\typeout{** the default language instead.}%
\else
\language=\csname l@#1\endcsname
\fi
#2}}
\providecommand{\BIBdecl}{\relax}
\BIBdecl

\bibitem[Rathore et~al.(2016)Rathore, Ahmad, Paul, and Rho]{rathore2016urban}
M.~M. Rathore, A.~Ahmad, A.~Paul, and S.~Rho, ``Urban planning and building smart cities based on the internet of things using big data analytics,'' \emph{Computer networks}, vol. 101, pp. 63--80, 2016.

\bibitem[Chen and Shang(2022)]{chen2022transformer}
G.~Chen and Y.~Shang, ``Transformer for tree counting in aerial images,'' \emph{Remote Sensing}, vol.~14, no.~3, p. 476, 2022.

\bibitem[Sarwar et~al.(2018)Sarwar, Griffin, Periasamy, Portas, and Law]{sarwar2018detecting}
F.~Sarwar, A.~Griffin, P.~Periasamy, K.~Portas, and J.~Law, ``Detecting and counting sheep with a convolutional neural network,'' in \emph{2018 15th IEEE International Conference on Advanced Video and Signal Based Surveillance (AVSS)}.\hskip 1em plus 0.5em minus 0.4em\relax IEEE, 2018, pp. 1--6.

\bibitem[Lei et~al.(2024)Lei, Xiao, Zhang, Li, Shi, and Zhu]{10816052}
S.~Lei, X.~Xiao, T.~Zhang, H.-C. Li, Z.~Shi, and Q.~Zhu, ``Exploring fine-grained image-text alignment for referring remote sensing image segmentation,'' \emph{IEEE Transactions on Geoscience and Remote Sensing}, pp. 1--1, 2024.

\bibitem[Liu et~al.(2024{\natexlab{a}})Liu, Tian, Zhao, Yu, Xie, Wang, Ye, and Liu]{liu2024vmambavisualstatespace}
\BIBentryALTinterwordspacing
Y.~Liu, Y.~Tian, Y.~Zhao, H.~Yu, L.~Xie, Y.~Wang, Q.~Ye, and Y.~Liu, ``Vmamba: Visual state space model,'' 2024. [Online]. Available: \url{https://arxiv.org/abs/2401.10166}
\BIBentrySTDinterwordspacing

\bibitem[Ma et~al.(2024)Ma, Lei, Celik, and Li]{ma2024fer}
H.~Ma, S.~Lei, T.~Celik, and H.-C. Li, ``{FER-YOLO-Mamba}: Facial expression detection and classification based on selective state space,'' \emph{arXiv preprint arXiv:2405.01828}, 2024.

\bibitem[Gu and Dao(2023)]{gu2023mamba}
A.~Gu and T.~Dao, ``Mamba: Linear-time sequence modeling with selective state spaces,'' \emph{arXiv preprint arXiv:2312.00752}, 2023.

\bibitem[Li et~al.(2018)Li, Zhang, and Chen]{li2018csrnet}
Y.~Li, X.~Zhang, and D.~Chen, ``{CSRNet}: Dilated convolutional neural networks for understanding the highly congested scenes,'' in \emph{2018 IEEE/CVF Conference on Computer Vision and Pattern Recognition}, 2018, pp. 1091--1100.

\bibitem[Wang et~al.(2019)Wang, Gao, Lin, and Yuan]{Wang2019sfcn}
Q.~Wang, J.~Gao, W.~Lin, and Y.~Yuan, ``Learning from synthetic data for crowd counting in the wild,'' in \emph{2019 IEEE/CVF Conference on Computer Vision and Pattern Recognition (CVPR)}, 2019, pp. 8190--8199.

\bibitem[Gao et~al.(2022)Gao, Liu, Hu, Li, Wen, and Wang]{gao2022psgcnet}
G.~Gao, Q.~Liu, Z.~Hu, L.~Li, Q.~Wen, and Y.~Wang, ``{PSGCNet}: A pyramidal scale and global context guided network for dense object counting in remote-sensing images,'' \emph{IEEE Transactions on Geoscience and Remote Sensing}, vol.~60, pp. 1--12, 2022.

\bibitem[Xu et~al.(2021)Xu, Liang, Zheng, Xie, and Ma]{xu2021dsaca}
W.~Xu, D.~Liang, Y.~Zheng, J.~Xie, and Z.~Ma, ``Dilated-scale-aware category-attention convnet for multi-class object counting,'' \emph{IEEE Signal Processing Letters}, vol.~28, pp. 1570--1574, 2021.

\bibitem[Gao et~al.(2024)Gao, Zhao, and Li]{gao2024moc}
J.~Gao, L.~Zhao, and X.~Li, ``{NWPU-MOC}: A benchmark for fine-grained multicategory object counting in aerial images,'' \emph{IEEE Transactions on Geoscience and Remote Sensing}, vol.~62, pp. 1--14, 2024.

\bibitem[Liu et~al.(2024{\natexlab{b}})Liu, Ma, Zhang, Wang, Ji, Sun, and Ji]{liu2024rotated}
S.~Liu, Y.~Ma, X.~Zhang, H.~Wang, J.~Ji, X.~Sun, and R.~Ji, ``Rotated multi-scale interaction network for referring remote sensing image segmentation,'' in \emph{Proceedings of the IEEE/CVF Conference on Computer Vision and Pattern Recognition}, 2024, pp. 26\,658--26\,668.

\bibitem[Guo et~al.(2024)Guo, Guo, Zha, Zhang, Li, Dai, Xia, and Li]{guo2024mambairv2}
H.~Guo, Y.~Guo, Y.~Zha, Y.~Zhang, W.~Li, T.~Dai, S.-T. Xia, and Y.~Li, ``{MambaIRv2}: Attentive state space restoration,'' \emph{arXiv preprint arXiv:2411.15269}, 2024.

\end{thebibliography}

\end{document}